\pdfoutput=1

\PassOptionsToPackage{x11names}{xcolor}
\documentclass[11pt]{article}

\usepackage[]{latex/acl}

\usepackage{times}
\usepackage{latexsym}
\usepackage{natbib}
\usepackage{kotex}
\usepackage{algorithm}
\usepackage{algorithmicx}
\usepackage[noend]{algpseudocode}
\usepackage{amssymb}
\usepackage{amsmath, amsfonts, amsthm}
\usepackage{dsfont}
\usepackage{mathtools}
\usepackage{amsthm}
\usepackage{booktabs} 
\usepackage{multirow}
\usepackage{subcaption}
\usepackage{adjustbox}
\usepackage{algpseudocode}
\usepackage[bottom, symbol]{footmisc}
\usepackage{hyperref}
\usepackage{empheq}
\usepackage[x11names]{xcolor}

\usepackage{soul} 

\newcommand{\ie}{\textit{i}.\textit{e}., }

\usepackage[T1]{fontenc}

\usepackage[utf8]{inputenc}

\usepackage{microtype}

\usepackage{inconsolata}

\newcommand\red[1]{\textcolor{Firebrick2}{#1}}
\newcommand\blue[1]{\textcolor{DodgerBlue2}{#1}}

\usepackage{soul}

\title{Mitigating Hallucination in Abstractive Summarization with Domain-Conditional Mutual Information}

\author{
  Kyubyung Chae\footnotemark[1] \;\; {\bf Jaepill Choi}\footnotemark[1] \;\; {\bf Yohan Jo} \;\; {\bf Taesup Kim}\footnotemark[2] \\
    Graduate School of Data Science, Seoul National University \\
  \texttt{\{kyubyung.chae, jaepill9205, yohan.jo, taesup.kim\}@snu.ac.kr}
}

\begin{document}
\maketitle

\footnotetext[1]{Equal Contribution.}
\footnotetext[2]{Corresponding author.}

\begin{abstract}


A primary challenge in abstractive summarization is hallucination---the phenomenon where a model generates plausible text that is absent in the source text. We hypothesize that the domain (or topic) of the source text triggers the model to generate text that is highly probable in the domain, neglecting the details of the source text. To alleviate this model bias, we introduce a decoding strategy based on domain-conditional pointwise mutual information.  This strategy adjusts the generation probability of each token by comparing it with the token's marginal probability within the domain of the source text. According to evaluation on the XSUM dataset, our method demonstrates improvement in terms of faithfulness and source relevance. The code is publicly available at \url{https://github.com/qqplot/dcpmi}.


\end{abstract}

\section{Introduction}
\label{sec:intro}


Abstractive summarization is the task of generating a summary by interpreting and rewriting a source text. State-of-the-art pre-trained language models have achieved remarkable performance in this task \cite{lewis2019bart, zhang2020pegasus}. However, upon closer examination, a common issue emerges: hallucination between the source document and the generated text. Prior studies have made efforts to enhance the faithfulness of the summary to the source text, yet hallucination remains a persistent challenge \cite{maynez-etal-2020-faithfulness, mao2020constrained, zhu-etal-2021-enhancing, zhang2023language}.

To solve this issue, we introduce a decoding strategy based on domain-conditional pointwise mutual information ($\text{PMI}_{\text{DC}}$). The motivation for $\text{PMI}_{\text{DC}}$ is that the domain of the source text provokes the model to generate text that is highly probable in the source domain, leading to plausible but factually inconsistent text. 
Building on this motivation, $\text{PMI}_{\text{DC}}$ computes how much more likely a token becomes in the summary when conditioned on the input source text, compared to when the token is conditioned only on the domain of the source text. This effectively penalizes the model's tendency to fall back to domain-associated words when the model has high uncertainty about the generated token.


\begin{table}[tb!]
\centering
\resizebox{0.95\columnwidth}{!}{%
\begin{tabular}{p{1.5cm}p{6.5cm}}
\toprule
\small\textbf{Method} & \small\textbf{Text} \\ 
\midrule

Source & { ...chairman of the Scottish Chambers of Commerce economic advisory group, said: ``Our latest economic data shows that many Scottish businesses \blue{\textbf{will have a successful 2017}}... } \\
\normalsize{CPMI} & {The Scottish Chambers of Commerce has issued a \red{\textbf{warning about the outlook for the economy in 2017.}}} \\
$\text{PMI}_{\text{DC}}$ &{The Scottish Chambers of Commerce has said it \blue{\textbf{expects the economy to have a ``successful'' year in 2017.}}} \\
\midrule
Domain & {Economy, Businesses, GDP}\\
\bottomrule
\end{tabular}%
}
\caption{An example of hallucination in abstractive summarization. Inconsistent words are highlighted in \red{\textbf{\textit{red}}} fonts, while consistent words are highligthed in \blue{\textbf{\textit{blue}}} fonts.}
\label{table:example}
\end{table}


This idea was inspired by conditional pointwise mutual information (CPMI) \cite{vanderpoel2022mutual}, which similarly penalizes a token's marginal probability. But CPMI does not capture the important fact that a token's probability depends highly on the source domain in summarization.
For example, consider the example presented in Table \ref{table:example}. The source text states, ``Our latest economic data shows that many Scottish businesses will have a successful 2017''.
CPMI undesirably introduces the term ``warning'', which frequently appears in the domain of economy in the training data, generating information that contradicts the source text. By contrast, $\text{PMI}_{\text{DC}}$ lowers the probability of the term ``warning'' by capturing the high conditional likelihood of this term given the domain and avoids the hallucination.

We use automated metrics for evaluation on the challenging XSUM dataset \cite{narayan-etal-2018-dont} achieving significant improvements in faithfulness and relevance to source texts according to metrics like AlignScore, FactCC, BARTScore, and BS-Fact, with only a marginal decrease in ROUGE and BERTScore. This highlights the effectiveness and robustness of $\text{PMI}_{\text{DC}}$ in abstractive summarization.

\begin{figure}[t]    
\centering
\includegraphics[width=0.9\columnwidth]{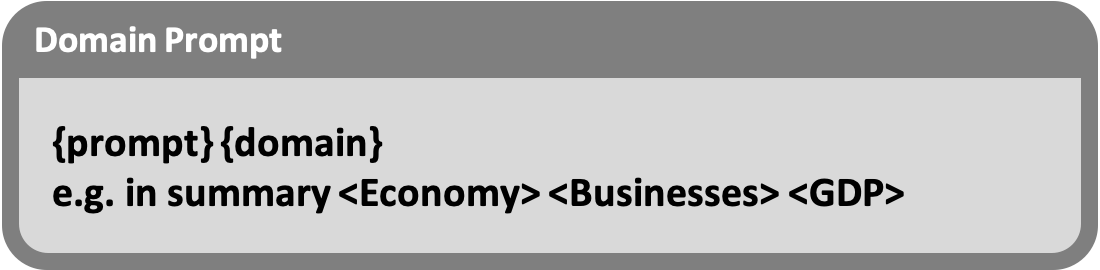}
\caption{Example of domain prompt.}     
\label{fig:domain_prompt}    
\end{figure}

\section{Preliminaries}
\label{sec:preliminaries}

\paragraph{Problem setting}
We adopt the problem definition in \citet{vanderpoel2022mutual}. In abstractive summarization, an input source text, denoted as $\mathbf{x} \in \mathcal{X}$, is condensed into an output string represented by $\mathbf{y} = \langle y_0, \ldots, y_T \rangle \in \mathcal{Y}$. This output string is a sequence of tokens from the vocabulary $\mathcal{V}$. Each sequence begins with token $y_0$ and ends with $y_T$, and the length of the output is $T+1$. 
The optimal $\mathbf{y}$ that belongs to a valid string set $\mathcal{Y}$ is obtained via a scoring function as follows:
$$
    \mathbf{y}^{*} = \underset{\mathbf{y} \in \mathcal{Y}}{\mathrm{argmax}}
    {\; \text{score}(\mathbf{y} | \mathbf{x})}.       
$$

Utilizing beam search is a practical solution for searching possible strings. 
The typical beam search with an autoregressive generation model uses the following scoring function:
\begin{equation}
    \text{score}(\mathbf{y} | \mathbf{x}) = \sum^{T}_{t=1}{\text{score} (y_t|\mathbf{x}, \mathbf{y}_{<t})}
    \label{eq:log_prob}
\end{equation}
where $\text{score}(y_t|\mathbf{x}, \mathbf{y}_{<t}) = \log p(y_t|\mathbf{x}, \mathbf{y}_{<t})$ is a token-level log probability computed by the model.


\paragraph{Pointwise Mutual Information}


PMI scoring utilizes mutual information between the input and output. This penalizes the generation of tokens that are marginally likely but not related to the input. The formula for PMI scoring can be expressed as follows:
\begin{equation}
\begin{aligned}
\mathrm{score}(y_t | \mathbf{x}, \mathbf{y}_{<t}) = 
&\log{p(y_t| \mathbf{x}, \mathbf{y}_{<t} )} \\
&- \log{p(y_t|\mathbf{y}_{<t})} 
\label{eq:pmi}
\end{aligned}
\end{equation}


\begin{table}[t]
\centering
\resizebox{1.0\columnwidth}{!}{%
\begin{tabular}{@{}c|lll@{}}
\toprule 
\textbf{Seed} & \textbf{Prompt Set} & {} & {}\\  
\midrule
\multirow{2}{*}{{keywords}} & \small{keywords} & \small{topics} & \small{components} \\
{} & \small{concepts} & \small{features} & \small{points} \\
\midrule
\multirow{2}{*}{{in summary}} & \small{in summary} & \small{to be brief} & \small{last of all} \\
{} & \small{when all is said and done} & \small{bringing up the rear} & \small{in short} \\
\midrule
\multirow{2}{*}{{in other words}} & \small{in other words} & \small{that is to say} & \small{to rephrase it} \\
{} & \small{take for example} & \small{to put it another way} & \small{case in point} \\
\bottomrule
\end{tabular}
}
\caption{Seed prompts and their corresponding paraphrased prompts. Each prompt was experimented to identify the most suitable prompts.}
\label{table:prompt_set}
\end{table}

\paragraph{Conditional Pointwise Mutual Information (CPMI)}

\citet{vanderpoel2022mutual} have demonstrated a connection between hallucinations and token-wise predictive entropy, denoted as $H(p) = -\sum_{y \in \mathcal{V}}{p_y \log{p_y}}$. A model tends to hallucinate a token if the entropy is high. Hence, instead of penalizing the marginal probability of $y_t$ in Equation~\ref{eq:pmi} all the time, CPMI does this only when the entropy at the $t$-th decoding step is higher than a threshold.
\begin{equation}
\begin{aligned}
\mathrm{score}(y_t | \mathbf{y}_{<t}, \mathbf{x}) = 
&\log{p_{\theta}(y_t|\mathbf{x}, \mathbf{y}_{<t}})\\
&- \lambda \cdot u_{t} \cdot \log{p_{\phi}(y_t| \mathbf{y}_{<t})} 
\end{aligned}
\label{eq:cpmi}
\end{equation}
where \(u_t = \mathds{1} \big\{ H\left(p_{\theta}(y_t |\mathbf{x}, y_{<t})\right) > \tau \big\} \). 





\begin{table*}[tb!]
\centering
\resizebox{0.9\textwidth}{!}{%
\begin{tabular}{@{}c|c|c|cccccc@{}}
\toprule 
\small{} &\small{} & \small{} 
& \multicolumn{2}{c}{\small{\textit{Faithfulness}}} & \multicolumn{2}{c}{\small{\textit{Relevance}}} & \multicolumn{2}{c}{\small{\textit{Similarity}}}\\ 
\small{\textbf{Method}} & \small{\textbf{Model}} & \small{\textbf{\# Samples}} & \small\textbf{AlignScore} & \small\textbf{FactCC} & \small\textbf{BARTScore$\uparrow$}  & \small\textbf{BS-Fact} & \small\textbf{ROUGE-L} & \small\textbf{BERTScore}\\ 
\midrule
\small{Beam}  & \multirow{4}{*}{\small{BART}}  & \small{11333} & \small{60.02} & \small{21.43} & \small{\underline{-1.8038}} & \small{\underline{88.86}} & \small{\underline{35.90}} & \small{\textbf{91.52}} \\
\small{PINOCCHIO}&  & \small{10647}\footnotemark[3] & \small{57.83} & \small{16.97} & \small{-2.0958} & \small{88.81} & \small{27.98} & \small{89.91}  \\
\small{CPMI} & & \small{11333} & \small{\underline{60.09}} & \small{\underline{21.53}} & \small{-1.8038} & \small{88.85} & \small{\textbf{35.90}} & \small{\underline{91.52}} \\
\small{PMI$_\text{DC}$} &  & \small{11333} & \small{\textbf{60.78$^{*}$}} & \small{\textbf{21.82}} & \small{\textbf{-1.7988$^{*}$}} & \small{\textbf{88.89$^{*}$}} & \small{35.81} & \small{91.50}\\
\bottomrule
\end{tabular}}
\caption{Comparison of different decoding methods on BART-large. PMI$_\text{DC}$ improves faithfulness and source relevance, with a slight decrease in target similarity. $*$ indicates statistical significance (p-value $< 0.001$) based on paired bootstrap analysis compared to CPMI.}
\label{table:main_result}
\end{table*}

\section{Domain-conditional Scoring Strategy}
\label{sec:method}

Our approach improves upon CPMI by conditioning the probability of a generated token on the source domain.
In our domain-conditional strategy ($\text{PMI}_{\text{DC}}$), we employ the following scoring function:

\begin{equation}
\begin{aligned}
\mathrm{score}(y_t | \mathbf{y}_{<t}
&, \mathbf{x}) =  \log{p_{\theta}(y_t|\mathbf{x}, \mathbf{x}_{dom}, \mathbf{y}_{<t}})\\
&- \lambda \cdot u_{t} \cdot \log{p_{\phi}(y_t|\mathbf{x}_{dom}, \mathbf{y}_{<t})} 
\end{aligned}
\label{eq:uncertainty-aware}
\end{equation}

 
$\mathbf{x}_{{dom}}$ is a domain prompt \cite{holtzman2022surface}, a subset of tokens in $\mathbf{x}$ that contains information about the source domain. This seemingly simple extension is well grounded in the previous observation that summarization models are likely to templatize the summaries of source texts that share the same domain or topic and hallucinate tokens that are frequent in the ``template'' of the source domain \cite{king2022dont}. Accordingly, our method can effectively account for different marginal probabilities of the token depending on the source domain and outperforms CPMI, as will be demonstrated later.

To compute the marginal probabilities $p(y_t | \mathbf{y}_{<t})$, we employ a smaller language model, denoted as $\phi$, while $\theta$ represents a larger summarization model. The hyperparameters $\lambda$ and $\tau$ are optimized through random grid-search.

\paragraph{Domain Prompt Design}

To condition the generation probability of a token on the source domain, we incorporate domain information into the prompts of both the summarization and language models (\ie $\mathbf{x}_{dom}$). We explored three types of domain information: (1) domain-specific keywords, (2) the first sentence of the source text, and (3) a randomly chosen sentence from the source text.




We assumed that domain-specific keywords enable the model to calculate the conditional probability of a token within the specified domain. The open-source module KeyBERT \cite{grootendorst2020keybert} was utilized to extract three keywords from each source text (Appendix \ref{subsec:keybert}). The expectation was that these selected keywords would effectively represent the source document with high similarity. Additionally, we also considered that sentences extracted from the source text could represent the domain of the entire text. Therefore, sentences from the source text, including the first sentence, and a randomly selected sentence were examined as the source domain.

\begin{table}[tb!]
\centering
\resizebox{\columnwidth}{!}{%
\begin{tabular}{@{}clcccccc@{}}
\toprule 
\small\textbf{Type} &\small\textbf{Domain}& \small\textbf{AlignScore}  & \small\textbf{BARTScore$\uparrow$}& \small\textbf{ROUGE-L}\\  
\midrule

\multirow{2}{*}{\small{Word}} & \small{Random} &   \small{60.47} & \small{-1.7993} & \small{35.82} \\
{} &   \small{Keyword} &   \small{60.78} & \small{-1.7988} & \small{35.81} \\
\midrule
\multirow{3}{*}{\small{Sentence}} &   \small{First} &  \small{61.45} & \small{-1.7706} & \small{35.52} \\

{} & \small{Random} &  \small{60.57} & \small{-1.7993} & \small{35.83} \\

{} & \small{Keyword}  & \small{61.16} & \small{-1.7784} & \small{35.60} \\ 
\bottomrule
\end{tabular}
}
\caption{Domain comparison. Results were obtained by varying the domain while using the BART model and the prompt ``\texttt{that is to say}.''}
\label{table:domain}
\end{table}
\footnotetext[3]{
For PINOCCHIO, we obtained results from 10,647 samples due to rejected paths. However the original paper reported results from 8,345 samples after manual removal. Thus, there may be discrepancies in our reported values.}

\begin{table}[t]
\centering
\resizebox{\columnwidth}{!}{%
\begin{tabular}{@{}c|c|cccc@{}}
\toprule 
\small\textbf{Method} &\small\textbf{FT}& \small\textbf{AlignScore}  & \small\textbf{BARTScore$\uparrow$}  & 
\small\textbf{ROUGE-L} \\  
\midrule
\small{Random} &   {} &  \small{\textbf{97.64}}  & \small{-2.6629} & \small{11.09}\\
\midrule
\small{FactPEG} &   \checkmark &  \small{68.70}  & \small{-1.9201}   & \small{34.36} \\
\small{$\text{PMI}_\text{DC}$}  & {}&  \small{60.78}   & \small{\textbf{-1.7988}} & \small{\textbf{35.81}} & \\
\bottomrule
\end{tabular}
}
\caption{Comparison with fine-tuned model. Random denotes the use of a randomly selected sentence from the source text as a summarization. FactPEG represents the summarization results obtained from a fine-tuned model with the objective of faithfulness.}
\label{table:finetune}
\end{table}

In conjunction with the aforementioned domain information, we incorporated a simple priming phrase into the domain prompt. We have discovered that using an appropriate lexical form yields better results compared to inputting the domain alone. We referred to the prompt design outlined by \citet{yuan2021bartscore}. The 18 phrases we examined include expressions such as ``\texttt{keyword},'' ``\texttt{in summary},'' and ``\texttt{in other words}.'' Table \ref{table:prompt_set} displays the seed prompts along with examples of paraphrased prompts (see more details in Appendix \ref{appendix:prompt_design}).


\section{Experimental Setup}
\label{sec:setup}

\paragraph{Dataset}
We used the eXtreme Summarization Dataset, XSUM \cite{narayan-etal-2018-dont}, which consists of BBC articles as source documents and single-sentence summaries as gold summaries. 

\paragraph{Baselines}

We examined three baseline decoding methods: standard beam search, PINOCCHIO \cite{king2022dont}, and CPMI \cite{vanderpoel2022mutual}. Additionally, we analyzed FactPEG \cite{wan2022factpegasus}, which underwent separate fine-tuning using FactCC and ROUGE with the source document.


\paragraph{Models}
For the summarization model, we utilized encoder-decoder structures of BART \cite{lewis2019bart} and PEGASUS \cite{zhang2020pegasus}. As for the language model, a GPT2-based model \cite{gpt2} was employed. Each of these models was pre-trained on the XSUM dataset. More details can be found in Appendix \ref{appendix:implement}.

\paragraph{Evaluation Metrics}
\label{subsec:metric}

We have categorized the evaluation into three key divisions: \textit{Faithfulness}, \textit{Relevance} (with the source), and \textit{Similarity} (with the target). For faithfulness, we used AlignScore \cite{zha2023alignscore} and FactCC \cite{kryscinski-etal-2020-evaluating}. To measure relevance to the source and informativeness, we employed BARTScore \cite{yuan2021bartscore} and BS-FACT. Lastly, to assess similarity to the target, we utilized ROUGE-L and BERTScore \cite{zhang2020bertscore}.

\begin{table}[t]
\centering
\resizebox{0.95\columnwidth}{!}{%
\begin{tabular}{p{1.5cm}p{6cm}}
\toprule
\small\textbf{Method} & \small\textbf{Text} \\ 
\midrule
FactPEG & The crypto-currency, Bitcoin.\\
$\text{PMI}_\text{DC}$ & The price of the virtual currency Bitcoin has fallen sharply in the wake of comments made by one of its most prominent developers.
\\
Source & Mike Hearn, a Zurich-based developer ...  published a blog calling Bitcoin a ``failed'' project ... Bitcoin's price fell quite sharply over the weekend ... \\
\bottomrule
\end{tabular}%
}
\caption{An example of FactPEG summary. The model trained with the objective of faithfulness tends to focus only on factual consistency, leading to a reduction in the summarization capability of pre-trained model.}
\label{table:FactPEGA vs ours}
\end{table} 

\section{Results}
\label{sec:results}

We presented the results from BART in Table \ref{table:main_result}. The complete result, including those from PEGASUS, are provided in Table \ref{table:main_appendix}. For all cases, the prompt used was ``\texttt{That is to say}'', and the domain consisted of three keywords extracted from the source document. In Table \ref{table:main_result}, we compared the summarization performance of different decoding strategies with BART. Our results revealed that PINOCCHIO exhibited suboptimal performance overall, while CPMI showed performance that was nearly on par with standard beam search. However, $\text{PMI}_\text{DC}$ showed significant improvement in terms of faithfulness and relevance. 

In Table \ref{table:domain}, the term \textit{Type} indicates whether the subset is at the word or sentence level, while \textit{Domain} refers to a subset of tokens within the source. Notably, the \textit{Keyword} approach within the word-level domain demonstrated robust performance. Therefore, we selected the \textit{Keyword} approach for our domain prompt.


\subsection{Comparison with Fine-tuned Model}

\begin{table}[tb!]
\centering
\resizebox{\columnwidth}{!}{%
\begin{tabular}{@{}c|ccccc@{}}
\toprule 
\small\textbf{Method}& \small\textbf{AlignScore}  & \small\textbf{BARTScore$\uparrow$}  & 
\small\textbf{ROUGE-L} \\  
\midrule
\small{PMI} &  \small{60.06}  & \small{-1.8041} & \small{\textbf{35.88}}\\
\small{$\text{PMI}_\text{DC}$ w/o ${\mathcal{\textit{u}_\textit{t}}}$} &  \small{60.57}  & \small{-1.7992}   & \small{35.76} \\
\small{$\text{PMI}_\text{DC}$ w/ ${\mathcal{\textit{u}_\textit{t}}}$} &  \small{\textbf{60.78}}   & \small{\textbf{-1.7988}} & \small{35.81} & \\
\bottomrule
\end{tabular}
}
\caption{Effectiveness of uncertainty-aware scoring. 
The first row indicates PMI scoring in Equation \ref{eq:pmi}. 
The second row denotes the removal of the uncertainty indicator (\ie $\textit{u}_\textit{t}$) from Equation \ref{eq:uncertainty-aware}. 
The third row refers to Equation \ref{eq:uncertainty-aware}. 
These results show the impact of the uncertainty indicator.}
\label{table:uncertainty-aware}
\end{table}

FactPEG \cite{wan2022factpegasus} reduces hallucinations by incorporating factual metrics during training, leveraging ROUGE and FactCC with the source document to produce faithful summaries. In Table \ref{table:finetune}, FactPEG outperforms {$\text{PMI}_\text{DC}$} in terms of faithfulness. On the other hand, {$\text{PMI}_\text{DC}$} achieves a more balanced performance across different metrics.

FactPEG is trained with a focus on faithfulness, which has led to the loss of other summarization abilities. For instance, using a random sentence as a summary (as shown in the top row in Table \ref{table:finetune}) demonstrates high faithfulness but a notable drop in the other two categories. Therefore, solely targeting faithfulness may risk the summarization capabilities of pre-trained models (refer to Table \ref{table:FactPEGA vs ours}).

\subsection{Effectiveness of Uncertainty-Aware Scoring}
\label{subsec:optimal_pmi}

Recall that in $\text{PMI}_{\text{DC}}$, the marginal probability of a token conditional to the domain $p(y_t|\mathbf{x}{_{dom}}, \mathbf{y}_{<t})$ is utilized only when the model's uncertainty of a token exceeds a threshold (\ie $u_t$). Here, we examined whether this uncertainty-aware scoring is more effective than without $u_t$.

In Table \ref{table:uncertainty-aware}, the first and second rows demonstrate the PMI scores regardless of uncertainty, while the third row shows uncertainty-aware PMI score. To ensure faithful token generation without degrading the performance of original summarization models, it is more effective to replace only specific uncertain tokens suspected of hallucination through uncertainty-aware scoring, rather than adjusting all tokens.

\subsection{Error Analysis}

While PMI$_\text{DC}$ effectively controlled hallucinated terms, there were instances of failure. We conducted a manual evaluation on 500 XSUM samples selected from \citet{maynez-etal-2020-faithfulness}, categorizing the error cases into three types (Table \ref{table:manual-evaluation}):

\begin{itemize}
\setlength\itemindent{-1em}
  \item \textbf{Case 1}: Extracted keywords may not fully reflect the domains of the source text. 
  \item \textbf{Case 2}: Appropriate domains, but errors in representing numbers, proper nouns, or statistics.
  \item \textbf{Case 3}: Appropriate domains, yet still hallucinated cases.
\end{itemize}

\begin{table}[t]
\centering
\resizebox{0.8\columnwidth}{!}{%
\begin{tabular}{@{}c|ccc@{}}
\toprule 
\small\textbf{Error case}& \small\textbf{\# of samples}  & \small\textbf{Percentage (\%)} \\  
\midrule
\small{Case 1} &  \small{120}  & \small{24.0}\\
\small{Case 2} &  \small{57}  & \small{11.4}\\
\small{Case 3} &  \small{55}   & \small{11.0} \\
\small{No error} &  \small{268}   & \small{53.6} \\
\midrule
\small{Total} &  \small{500}   & \small{100.0} \\
\bottomrule
\end{tabular}
}
\caption{Manual evaluation on 500 XSUM samples. Initially, samples with an AlignScore of 0.5 or lower were considered potential error cases. Subsequently, two co-authors annotated each potential error sample, categorizing them as Case 1, Case 2, Case 3, or No error.}
\label{table:manual-evaluation}
\end{table}

\textbf{Case 1} occurs when the extracted keywords may not fully reflect the domains of the source text. We used keywords to represent the domain. However, in some cases, the extracted keywords may not adequately capture the ``topic'' or ``category'' of the source text and did not guide the model as we expected (Table \ref{table:error1}). 

\textbf{Case 2} occurs when handling numbers, proper nouns, or statistics. Numbers, proper nouns, or statistics are among the primary causes of hallucination in the model. Despite extracting the appropriate domain, there are instances where incorrect numerical information is presented in the generated text (Table \ref{table:error2}).

\textbf{Case 3} refers to situations where summarization fails even though they do not fall into Case 1 or Case 2. One such scenario happens when imposing significant penalties on domain-specific keywords. This can result in avoiding direct expressions, leading to ambiguity (Table \ref{table:error3}). Additionally, there are occurrences of hallucination due to the inherent difficulty of the task. For instance, when the source text contains multiple pieces of information, summarizing them into a single sentence becomes a challenging task. 

\section{Conclusion}
\label{sec:conclusion}


We proposed a decoding strategy based on domain-conditional pointwise mutual information (PMI$_\text{DC}$) to reduce hallucination in abstractive summarization. PMI$_\text{DC}$ penalizes the model's tendency to generate text inconsistent with the source document by considering the source text's domain. This simple but innovative approach significantly improves faithfulness and relevance to the source text, as demonstrated through evaluation on the XSUM dataset.





\section*{Limitations}
\label{sec:limitation}

While our method demonstrated improvements in faithfulness and source relevance with BART and PEGASUS on the XSUM dataset, these enhancements are relatively modest across the board. Further exploration and validation are needed, especially through experimentation with other models and diverse datasets to evaluate their efficacy under varied conditions.

Additionally, our evaluation process has limitations, as comprehensive human evaluations across the entire dataset were not conducted. Human evaluation remains the most reliable measure for assessing hallucinations in summarization tasks, providing insights that automated metrics may lack. However, given that human evaluation can also be influenced by biases and subjectivity \cite{maynez-etal-2020-faithfulness}, future research should integrate more extensive human evaluations alongside automated assessments to provide a more comprehensive evaluation of model performance.

\section*{Ethical Concerns} 
We do not anticipate any ethical concerns with this work beyond those already documented in abstractive summarization systems and other text generators \cite{vanderpoel2022mutual, zhou2023contextfaithful, xiao2021hallucination}.

\section*{Acknowledgements} 
This research was supported by the National Research Foundation of Korea (NRF) grant funded by the Korea government (MSIT) (No. RS-2023-00222663). We would like to thank Gwangho Choi for his valuable discussions.

\bibstyle{latex/acl_natbib.bst}
\bibliography{custom}

\appendix

\clearpage
\section{Related Work}
\label{sec:related_work}

\subsection{Understanding hallucinations}




In abstractive summarization, \textit{hallucinations} occur when the generated content diverges from the source material, categorized as intrinsic and extrinsic hallucinations \cite{maynez-etal-2020-faithfulness}. Intrinsic hallucinations arise from generating content that contradicts the input source document, while extrinsic hallucinations occur from ignoring the source \cite{Ji_2023}. Our focus lies in summarization, where a quality summary mirrors the source's content. Thus, reducing hallucinations entails increasing \textit{faithfulness} and \textit{factual consistency} between the source document and the generated summary.


\cite{zhang2023language} highlighted the snowball effect of hallucination: initial inaccuracies tend to propagate subsequent incorrect explanations due to \textit{initial commitment}. Language models, trained on data where the correct answer precedes the explanation, tend to align subsequent explanations with initial inaccuracies. Hence, early correction of hallucinated content is crucial.

\subsection{Mitigating hallucinations}
Various approaches have been proposed to tackle the challenge of hallucination in text generation \cite{li2022faithfulness}.

Lexically constrained decoding modifies beam search to control specific words in the output without changing the model. Constrained Abstractive Summarization (CAS) \cite{mao2020constrained} uses dynamic beam search to create constrained token sets, improving the accuracy and faithfulness of abstractive summarization.

PINOCCHIO \cite{king2022dont} is a modified beam search algorithm utilizing a rejected set $\mathcal{R}$ to avoid disallowed paths. It tackles inconsistencies by adjusting predicted scores and employing backtracking with a heuristic function $f_c$, which incorporates eight binary checks. Thus generations with high entropy and multiple backtracks are discarded.

Context-aware decoding (CAD) \cite{shi2023trusting} attempts to decrease hallucination by adding prompts to the unconditional term in PMI. However, unlike our method, CAD adjusts the score of all tokens and applies the same prompt for all input documents.

CPMI \cite{vanderpoel2022mutual}, a significant inspiration for our work, introduced a beam search technique to address hallucination. It tackles the tendency of language models to produce overly general text by utilizing mutual information and internal entropy in a scoring function, thus detecting and mitigating hallucination.

Additionaly, \citet{xiao2021hallucination} introduced an uncertainty-aware beam search method that penalizes the usage of entropy. In contrast, our approach diverges by not consistently penalizing uncertain tokens; instead, we score them with PMI when they exceed a specific threshold.

FactPegasus \cite{wan2022factpegasus} enhances abstractive summarization by reducing hallucinations through factuality integration. It modifies sentence selection by combining ROUGE and FactCC, aiming for faithful summaries. FactPegasus employs fine-tuning with \textit{corrector}, \textit{contrastor}, and \textit{connector} modules. Although it improves factual consistency, it lacks in informativeness. Our work proposes a more balanced abstractive summarization approach.

\subsection{Automatic Metrics}
We have categorized the evaluation metrics into three key dimensions: \textit{Faithfulness}, \textit{Relevance} (with the source), and \textit{Similarity} (with the target). 

To assess faithfulness, we employed \textbf{AlignScore} \cite{zha2023alignscore} and \textbf{FactCC} \cite{kryscinski-etal-2020-evaluating}. AlignScore divides the source document into approximately 350 segments, evaluating factual consistency with the generated text. FactCC assesses whether the generated text aligns factually with the source document, using a binary format.

To evaluate the relevance of the generated text with the source document, we used \textbf{BARTScore} \cite{yuan2021bartscore} and \textbf{BS-FACT}. BARTScore, which is based on the BART model, comprehensively evaluates both the informativeness and factual accuracy of the generated text. BS-FACT, derived from BERTScore, measures the precision of alignment between the generated text and the source text.

Finally, to measure similarity with the target, we utilized \textbf{ROUGE-L} \cite{lin-2004-rouge} and \textbf{BERTScore} \cite{zhang2020bertscore}. These metrics, traditionally used for evaluating generated text, differ from previous methods as they compare the generated text not with the source document but with the gold summary (\textit{\ie target}).

\begin{table*}[tb!]
\centering
\resizebox{0.9\textwidth}{!}{%
\begin{tabular}{@{}c|c|c|cccccc@{}}
\toprule 
\small{} &\small{} & \small{} 
& \multicolumn{2}{c}{\small{\textit{Faithfulness}}} & \multicolumn{2}{c}{\small{\textit{Relevance}}} & \multicolumn{2}{c}{\small{\textit{Similarity}}}\\ 
\small{\textbf{Method}} & \small{\textbf{Model}} & \small{\textbf{\# Samples}} & \small\textbf{AlignScore} & \small\textbf{FactCC} & \small\textbf{BARTScore$\uparrow$}  & \small\textbf{BS-Fact} & \small\textbf{ROUGE-L} & \small\textbf{BERTScore}\\ 
\midrule
\small{Beam}  & \multirow{4}{*}{\small{BART}}  & \small{11333} & \small{60.02} & \small{21.43} & \small{\underline{-1.8038}} & \small{\underline{88.86}} & \small{\underline{35.90}} & \small{\textbf{91.52}} \\
\small{PINOCCHIO}&  & \small{10647} & \small{57.83} & \small{16.97} & \small{-2.0958} & \small{88.81} & \small{27.98} & \small{89.91}  \\
\small{CPMI} & & \small{11333} & \small{\underline{60.09}} & \small{\underline{21.53}} & \small{-1.8038} & \small{88.85} & \small{\textbf{35.90}} & \small{\underline{91.52}} \\
\small{PMI$_\text{DC}$} &  & \small{11333} & \small{\textbf{60.78}} & \small{\textbf{21.82}} & \small{\textbf{-1.7988}} & \small{\textbf{88.89}} & \small{35.81} & \small{91.50}\\

\midrule
\small{Beam}  & \multirow{3}{*}{\small{PEGASUS}}  & \small{11333} 
& \small{59.28} & \small\underline{22.02} & \small{{-1.9636}} & \small{{88.64}} & \small{{38.02}} & \small{\textbf{91.91}} \\
\small{CPMI} & & \small{11333} & \small{\underline{59.31}} & \small{{21.91}} & \small\underline{{-1.9617}} & \small\underline{{88.64}} & \small{\underline{38.01}} & \small{\underline{91.91}} \\
\small{PMI$_\text{DC}$} &  & \small{11333} & \small{\textbf{59.40}} & \small{\textbf{22.09}} & \small{\textbf{-1.9590}} & \small{\textbf{88.64}} & \small{\textbf{38.06}} & \small{91.91}\\
\bottomrule
\end{tabular}}
\caption{Comparison with decoding methods on BART-large and PEGASUS. PMI$_\text{DC}$ improves faithfulness and source relevance, with a slight decrease in target similarity.}
\label{table:main_appendix}
\end{table*}

\subsection{Keyword Extractor}
\label{subsec:keybert}


We utilized the open-source module KeyBERT \cite{grootendorst2020keybert} to extract keywords from the source document. KeyBERT utilizes \texttt{all-MiniLM-L6-v2} model, a sentence-transformers model designed to map sentences and paragraphs into a 384-dimensional dense vector space, facilitating tasks like clustering or semantic search. This model is based on the pre-trained model \texttt{nreimers/MiniLM-L6-H384-uncased}, fine-tuned on over 1 billion sentence pairs using a contrastive learning objective. It is specifically modeled for encoding sentences and short paragraphs, thus enabling the generation of semantic vectors for tasks like information retrieval, clustering, or assessing sentence similarity (\url{https://huggingface.co/sentence-transformers/all-MiniLM-L6-v2}).

\section{Implementation Details}
\label{appendix:implement}

\paragraph{Summarization models} 
In our experiments, we followed a setup akin to that described in \citet{vanderpoel2022mutual} to ensure a fair comparison. Our experiments were conducted on computing clusters equipped with NVIDIA RTX 3090 GPUs, allocating a single GPU for each experiment. We utilized the BART-large-XSUM checkpoint (\url{https://huggingface.co/facebook/bart-large-xsum}) and the PEGASUS-XSUM checkpoint (\url{https://huggingface.co/google/pegasus-xsum}).

\paragraph{Language model}
We trained two language models, one for BART-large and one for PEGASUS. Both language models belong to the GPT2 family \cite{gpt2} (available at \url{https://huggingface.co/gpt2}). The configurations for the language models are identical: 512 embeddings, 6 layers, and 8 heads. However, there is a discrepancy in the output vocabulary size, with BART at 50,265 and PEGASUS at 96,103. Both models have a maximum token length set to 2,048 tokens, and operate with an update frequency of 32. They share a learning rate of $5.0 \times 10^{-4}$. For validation metrics, BART-large consisted a loss of 3.16744 and a perplexity of 24.57401, while PEGASUS consisted a loss of 3.25238 and a perplexity of 26.68345.

\paragraph{Why do we need an additional model?}
We have employed two types of models: a larger summarization model (BART-large: 406M, PEGASUS: 223M) and a smaller language model (GPT2-based model: 45M). There are two reasons why we chose to use an additional decoder-only language model instead of reusing the decoder of the summarization model.

First of all, an extra forward pass is required for the unconditional (\ie domain-conditional) term. Therefore, employing a smaller language model is faster. This aligns with recent research on speeding up additional forwarding, such as speculative sampling techniques \cite{chen2023accelerating}.

Secondly, a decoder-only structure, trained for the next token prediction, provides a more suitable unconditional distribution than an encoder-decoder structure. In an encoder-decoder architecture, the decoder relies on encoder output for cross-attention. Therefore, despite padding all encoder inputs, an appropriate unconditional distribution isn't achieved due to some samples lacking a source document in the training dataset.

\section{Searching Hyperparameters}
We adopted the hyperparameters reported in the CPMI paper for consistency. For BART, we configured $\tau$ to $3.5987$ and $\lambda$ to $6.5602 \times 10^{-2}$. Our approach surpassed CPMI's performance, demonstrating effective summarization without hallucination (refer to Table \ref{table:main_appendix}). For PEGASUS, we determined the hyperparameters by examining the AlignScore with 3,000 samples from the validation set, using CPMI, not $\text{PMI}_{\text{DC}}$. The values we obtained are $\tau = 3.304358$ and $\lambda = 7.4534 \times {10}^{-2}$. Note that CPMI relied on human-annotated data at the token level \cite{zhou21aclfindings}. This method is not only extremely costly and challenging but also lacks precision. However, since we have eliminated such human intervention, $\text{PMI}_\text{DC}$ is more applicable. 

\begin{figure}[t]    
    \begin{subfigure}{\columnwidth}
        \includegraphics[width=\columnwidth]{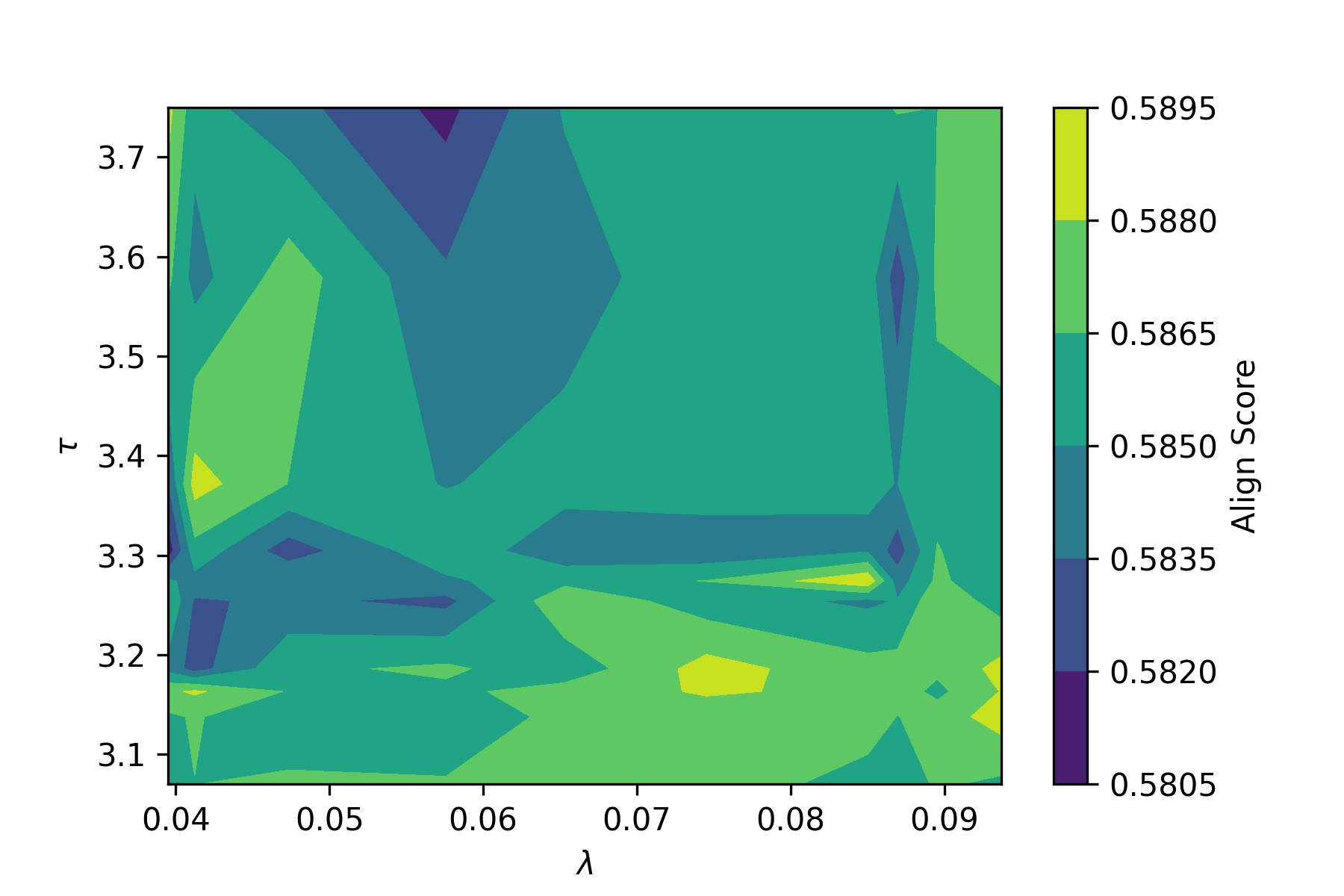}
        \caption{PEGASUS. CPMI. AlignScore}
    \end{subfigure}
    \quad 
    \begin{subfigure}{\columnwidth}
        \includegraphics[width=\columnwidth]{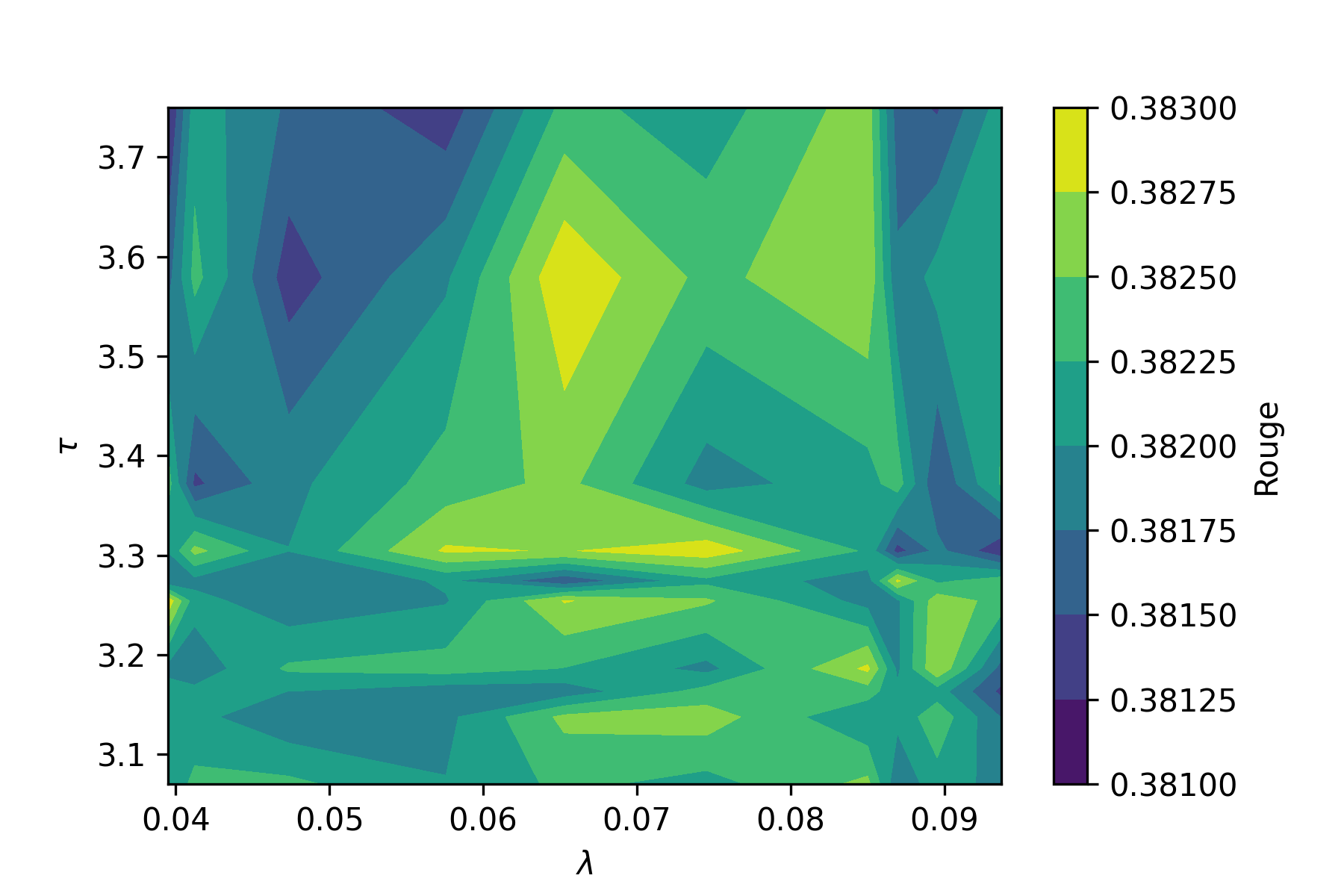}
        \caption{PEGASUS. CPMI. ROUGE-L}
    \end{subfigure} 
\caption{Hyperparameter search for PEGASUS. To ensure comparability with CPMI, identical hyperparameter settings were employed. A random uniform grid search was performed on 3,000 samples from a validation set, considering 10$\times$10 hyperparameter pairs based on AlignScore. Alternatively, optimization based on ROUGE-L scores was also explored, indicating that the optimal configuration may differ based on experimental outcomes.}        
\label{fig:hyperparameters}    
\end{figure}

\section{Prompt Design}
\label{appendix:prompt_design}

In our search for the best prompt, we referred to the prompt set proposed by \citet{yuan2021bartscore}. Their approach involved manually crafting seed prompts and collecting paraphrases to construct the prompt set, with the aim of finding suitable prompts within a defined search space. 

The average results presented in Table \ref{table:prompt_results} incorporate 19 prompts, including scenarios where no prompt is used. These results consistently demonstrate superior performance in faithfulness metrics compared to CPMI, highlighting the importance of \texttt{domain} information. The rationale behind prepending prompts to the domain is to seamlessly integrate domain information without deteriorating the naturalness of the language model. Our findings suggest that augmenting prompts is more effective than using domain alone.

\begin{table*}[tb!]
\centering
\resizebox{0.9\textwidth}{!}{%
\begin{tabular}{lcccccc}
\toprule
Prompt                    & AlignScore & FactCC  & BARTscore$\uparrow$ & BS-FACT  & ROUGE-L  & BERTscore\\
\midrule
w/o                       & 60.48    & 22.02           & -1.8033 & 88.88       & 35.81 & 91.50\\
\midrule
keywords                  & 60.45    & 21.94           & -1.8039 & 88.88       & 35.76 & 91.49\\
topics                    & 60.40    & 21.63           & -1.8063 & 88.88       & 35.78 & 91.50\\
components                & 60.67    & 21.72           & -1.8036 & 88.88       & 35.76 & 91.50\\
concepts                  & 60.48    & 21.76           & -1.8047 & 88.88       & 35.81 & 91.51\\
features                  & 60.53    & 21.66           & -1.8041 & 88.88       & 35.81 & 91.50\\
points                    & 60.37    & 21.57           & -1.8088 & 88.87       & 35.79 & 91.50\\
\midrule
in summary                & 60.55    & 21.67           & -1.8052 & 88.88       & 35.70 & 91.49\\
to be brief               & 60.33    & 21.58           & -1.8032 & 88.88       & 35.81 & 91.50\\
last of all               & 60.42    & 21.53           & -1.8035 & 88.88       & 35.80 & 91.50\\
when all is said and done & 60.66    & 21.59           & -1.8012 & 88.88       & 35.75 & 91.50\\
bringing up the rear      & 60.64    & 21.68          & -1.8020 & 88.89       & 35.80 & 91.50\\
in short                  & 60.67    & 21.63          & -1.8035 & 88.88       & 35.78 & 91.51\\
\midrule
in other words            & 60.71    & 21.71          & -1.7988 & 88.88       & 35.80 & 91.51\\
that is to say            & 60.78    & 21.82          & -1.7988 & 88.89       & 35.81 & 91.50\\
to rephrase it            & 60.66    & 21.96          & -1.8011 & 88.89       & 35.80 & 91.50\\
take for example          & 60.76    & 21.87          & -1.8025 & 88.88       & 35.81 & 91.50\\
to put it another way     & 60.45    & 21.69          & -1.8013 & 88.89       & 35.76 & 91.49\\
case in point             & 60.62    & 21.81          & -1.8033 & 88.87      & 35.81 & 91.51\\
\bottomrule
\end{tabular}}
\caption{Results for each prompt, where the \texttt{domain} consists of three keywords. Adding ``\texttt{that is to say}'' to the three keywords yielded the best overall performance.}
\label{table:prompt_results}
\end{table*}

\begin{table*}[tb!]
\centering
\resizebox{\textwidth}{!}{%
\begin{tabular}{p{1.5cm}p{15cm}}
\toprule
\small\textbf{Method} & \small\textbf{Text} \\ 
\midrule
Domain & {bia, falkirk, bi} \\
\midrule
Source & {However, the Bairns boss has underlined that any forward signing will need to exhibit even more quality than two of his promising youngsters. ``If I bring another striker in he's got to be better than young Botti \textbf{Bia}-\textbf{Bi} and Scott Shepherd,'' said Houston. ``I would be looking for the more experienced type, and another defender would come in handy as well.'' Eighteen-year-old Bia-Bi, a London-born Scot who has progressed through \textbf{Falkirk}'s academy, glanced in a fine equalising header against Cowdenbeath on Saturday to ensure Houston's side left Central Park with a point...} \\
\midrule
$\text{PMI}_{\text{DC}}$ & {Falkirk manager Peter Houston has not ruled out bringing in a new striker in the \red{January transfer window}. } \\
\normalsize{Gold} & {Peter Houston is still seeking to fine-tune his Falkirk squad, with a striker and defender pinpointed as priorities.} \\
\bottomrule
\end{tabular}%
}
\caption{\textbf{Case 1 error.} Inconsistent words are highlighted in \red{\textit{red}} fonts. Extracted keywords may not fully reflect domains of source text. In this example, the domain should be more related to terms like \textit{transfer} or \textit{football} rather than specific names of individuals or institutions. Hence, terms closely related with transfer (such as \textit{January}) were not adequately penalized.}
\label{table:error1}
\end{table*}

\begin{table*}[tb!]
\centering
\resizebox{\textwidth}{!}{%
\begin{tabular}{p{1.5cm}p{15cm}}
\toprule
\small\textbf{Method} & \small\textbf{Text} \\ 
\midrule
Domain & {invest, richest, investment} \\
\midrule
Source & {The \textbf{investment} follows ``several months of negotiations'', a company statement to the Saudi stock exchange said. The prince, who is one of the world's \textbf{richest} men, owns stakes in many well-known companies, including News Corporation. He also has investments in a number of media groups in the Arab world. ``Our investment in Twitter reaffirms our ability in identifying suitable opportunities to \textbf{invest} in promising, high-growth businesses with a global impact.'' Prince Alwaleed said.} \\
\midrule
$\text{PMI}_{\text{DC}}$ &{Saudi Arabia's Prince Alwaleed bin Talal has bought a \red{10\%} stake in Twitter \red{in a deal worth \$2bn (31.8bn)}.
} \\
\normalsize{Beam} & {Saudi Arabia's Prince Alwaleed bin Talal has agreed to buy a \red{10\%} stake in Twitter \red{for \$3bn (32.3bn)}.} \\
\bottomrule
\end{tabular}%
}
\caption{\textbf{Case 2 error.} Inconsistent words are highlighted in \red{\textit{red}} fonts. The appropriate domain, but not properly regulated in accounting numbers. Hallucinations related to proper nouns, numbers and statistics, have long been significant issues in language models. Our approach could not completely address this issue.}
\label{table:error2}
\end{table*}

\begin{table*}[tb!]
\centering
\resizebox{\textwidth}{!}{%
\begin{tabular}{p{1.5cm}p{15cm}}
\toprule
\small\textbf{Method} & \small\textbf{Text} \\ 
\midrule
Domain & {claire, marathon, equestrian} \\
\midrule
Source & {When \textbf{Claire} was told she would spend the rest of her life in a wheelchair after a spinal injury, she wanted to get back on her feet as quickly as possible and regain her independence. For the past three months she has been training intensively for the \textbf{marathon} using a robotic walking suit to prove she is just as determined as in her sporting days.  
... former champion British \textbf{equestrian} Lucinda Green. ``There's a lot of people who are worse off than me and haven't got the support I've got, so I want to raise as much as I can.'' But, when the marathon is over, Claire thinks that for the first time in six years, she will be delighted to return to her wheelchair.} \\
\midrule
$\text{PMI}_{\text{DC}}$ &{A paralysed equestrian rider is taking part in the London Marathon in a bid to become \red{the first person} in the world to walk unaided.
} \\
\normalsize{Beam} & {Claire \red{Gwynne}, who was paralysed from the chest down in 2006, is taking part in the London Marathon.} \\
\bottomrule
\end{tabular}%
}
\caption{\textbf{Case 3 error.} Inconsistent words are highlighted in \red{\textit{red}} fonts. Constraints of domain-conditional term can prevent direct expressions, potentially leading to ambiguity and generating incorrect results. 
In this example, penalizing the domain term \textit{Claire} led to the removal of the hallucinated term \textit{Gwynne}. However, beyond this correction, the conveyed information remained somewhat inaccurate.}
\label{table:error3}
\end{table*}




\end{document}